\pdfoutput=1

\documentclass[11pt]{article}

\usepackage{EMNLP2023}

\usepackage{times}
\usepackage{latexsym}

\usepackage[T1]{fontenc}

\usepackage[utf8]{inputenc}

\usepackage{microtype}

\usepackage{inconsolata}

\newcommand{\overentity}[2]{$\overbrace{\textsf{#1}}^\textsf{\textcolor{magenta}{#2}}$}
\newcommand{\underentity}[2]{$\underbrace{\textsf{#1}}_\textsf{\textcolor{magenta}{#2}}$}

\usepackage{algorithm}
\usepackage{algorithmic}

\usepackage{booktabs}
\usepackage{multirow}

\usepackage{caption}
\usepackage{subcaption}

\usepackage{enumitem}
\setlist{noitemsep}

\usepackage{tikz}
\usetikzlibrary{positioning}

\tikzset{
    textbox/.style={rectangle, rounded corners, draw, inner sep=8pt},
    textbox_input/.style={textbox, fill=blue!10},
    textbox_output/.style={textbox, fill=green!10},
    textbox_prompt/.style={textbox, fill=gray!10},
    textbox_generation/.style={textbox, fill=yellow!10},
}

\newcommand{\egseqmessage}{
    Hi, Can we please amend the following POs: \overentity{4505858}{Order Number} - increase \overentity{RDHRJIBJRGIJN}{Product ID} by \overentity{83}{Quantity}, \overentity{2077888476}{Order Number} - increase \underentity{LUCQHRSMCOVMM}{Product ID} by \underentity{53}{Quantity}. Thanks!\\[8pt]
    \textcolor{blue}{\textbf{Intent}: Order > Amendment > Remove}
}

\newcommand{\egsubject}{Orders}
\newcommand{\egbody}{%
    Please can we:\\[2pt]
    Increase 78WTH6 by 379 and remove W8E9HW4YN \& 138465B312I from order 25N246NO2.\\[2pt]
    And also remove 09W84YH from 29485H7NYO24N.\\[2pt]
    Thanks.
}

\newcommand{\egmessage}{
    \textbf{Subject:} \egsubject\\[2pt]
    \textbf{Message:} \egbody
}

\newcommand{\egintents}{
    Order > Amendment > Increase Quantity By\\[2pt]
    Order > Amendment > Remove\\
}

\newcommand{\egentities}{
    Order > Amendment > Increase Quantity By:
    \begin{itemize}
        \vspace{-6pt}
        \item Order Number
        \item Product ID
        \item Quantity
    \end{itemize}
    Order > Amendment > Remove:
    \begin{itemize}
        \vspace{-6pt}
        \item Order Number
        \item Product ID
    \end{itemize}
}

\newcommand{\egextraction}{
    ```[\\
    \quad \{\\
        \qquad "intent": "Order > Amendment > Increase Quantity By",\\
        \qquad "order\_number": "25N246NO2",\\
        \qquad "product\_id": "78WTH6",\\
        \qquad "quantity": "379"\\
    \quad \},\\
    \quad \{\\
        \qquad "intent": "Order > Amendment > Remove Item",\\
        \qquad "order\_number": "25N246NO2",\\
        \qquad "product\_id": "W8E9HW4YN"\\
    \quad \},\\
    \quad \{\\
        \qquad "intent": "Order > Amendment > Remove Item",\\
        \qquad "order\_number": "25N246NO2",\\
        \qquad "product\_id": "138465B312I"\\
    \quad \},\\
    \quad \{\\
        \qquad "intent": "Order > Amendment > Remove Item",\\
        \qquad "order\_number": "29485H7NYO24N",\\
        \qquad "product\_id": "09W84YH"\\
    \quad \}\\
    ]'''\\
}

\newcommand{\egsubjectoneshot}{Need more of these two items}
\newcommand{\egbodyoneshot}{%
    Hey,\\[2pt]
    The order for these two items didn't come in as expected. Can you increase the quantity for the following items?\\[2pt]
    BCZWSUANAFZEDV - order number 1427338571 - increase by 73\\[2pt]
    72435672357161 - order number MCR5VPPU - increase by 45\\[2pt]
    Thanks,\\[2pt]
    Kim
}

\newcommand{\egextractiononeshot}{
    [\{\\
        \quad "intent": "Order > Amendment > Increase Quantity By",\\
        \quad "order\_number": "1427338571",\\
        \quad "product\_id": "BCZWSUANAFZEDV",\\
        \quad "quantity": "73"\\
    \},\\
    \{\\
        \quad "intent": "Order > Amendment > Increase Quantity By",\\
        \quad "order\_number": "MCR5VPPU",\\
        \quad "product\_id": "72435672357161",\\
        \quad "quantity": "45"\\
    \}]
}

\newcommand{\egpromptoneshot}{
    \textcolor{purple}{Write a JSON summarizing the text.\\[8pt]
    The valid JSON schemas are:\\[2pt]
    \{"intent": "Order > Amendment > Increase Quantity By", "order\_number": "", "product\_id": "", "quantity": ""\}\\[8pt]
    {[[Text:]]}\\[2pt]
    Subject: \egsubjectoneshot\\[2pt]
    Message: \egbodyoneshot\\[8pt]
    {[[JSON:]]}\\
    \egextractiononeshot\\[8pt]}
    Write a JSON summarizing the text.\\[8pt]
    The valid JSON schemas are:\\[2pt]
    \{"intent": "Order > Amendment > Increase Quantity By", "order\_number": "", "product\_id": "", "quantity": ""\}\\[2pt]
    \{"intent": "Order > Amendment > Remove Item", "order\_number": "", "product\_id": ""\}\\[8pt]
    {[[Text:]]}\\[2pt]
    Subject: \egsubject\\[2pt]
    Message: \egbody\\[8pt]
    {[[JSON:]]}\\
    {[\{}
}

\newcommand{\egwikifull}{
    The Law Brook or Postford Brook is a stream in the Surrey Hills AONB which feeds the Tillingbourne which in turn feeds the River Wey.\\[2pt]
    Albury is a village and civil parish in the borough of Guildford in Surrey, England, about 4 miles (6.4 km) south-east of Guildford town centre.\\[2pt]
    The River Wey is a tributary of the River Thames in south east England. Its two branches, one of which rises near Alton in Hampshire and the other in West Sussex to the south of Haslemere, join at Tilford in Surrey.\\[2pt]
    The Surrey Hills is a 422 km2 (163 sq mi) Area of Outstanding Natural Beauty (AONB), which principally covers parts of the North Downs and Greensand Ridge in Surrey, England (approximately one quarter of the land area of the county).\\[2pt]
}

\newcommand{\egwikitypes}{
    River\\[2pt]
    Settlement\\[2pt]
    ProtectedArea
}

\newcommand{\egwikientities}{
    River:
    \begin{itemize}
        \vspace{-6pt}
        \item Name
        \item Subdivision Name
        \item Mouth
        \item Mouth Location
        \item Source 1 Location
    \end{itemize}
    Settlement:
    \begin{itemize}
        \vspace{-6pt}
        \item Name
        \item Official Name
        \item Subdivision Name
    \end{itemize}
    ProtectedArea:
    \begin{itemize}
        \vspace{-6pt}
        \item Name
        \item Location
        \item Established
    \end{itemize}
}

\newcommand{\egwikifullextraction}{
    [\\
    \quad \{\\
        \qquad "type": "River",\\
        \qquad "name": "Law Brook",\\
        \qquad "subdivision\_name": "Surrey"\\
    \quad \},\\
    \quad \{\\
        \qquad "type": "Settlement",\\
        \qquad "name": "Albury"\\
    \quad \},\\
    \quad \{\\
        \qquad "type": "River",\\
        \qquad "name": "River Wey"\\
        \qquad "subdivision\_name": "England",\\
        \qquad "mouth": "River Thames",\\
        \qquad "mouth\_location": "Surrey",\\
        \qquad "source1\_location": "Hampshire",\\
    \quad \},\\
    \quad \{\\
        \qquad "type": "ProtectedArea",\\
        \qquad "name": "Surrey Hills",\\
        \qquad "location": "Surrey, England"\\
    \quad \}\\
    ]
}

\newcommand{\eggptprompt}{
    Write an email explaining what the user wants to do, based on the JSON. Include all of the values from the JSON.\\[8pt]
    {[[JSON:]]}\\[2pt]
    [\{"intent": "Order > Cancel", "order\_number": "4ZM7QGRT"\}, \{"intent": "Return > Label", "return\_id": "77436508"\}]\\[8pt]
    {[[Email:]]}\\[2pt]
    Subject: Cancel order and returns label\\[2pt]
    Message: I made two orders a month ago (4ZM7QGRT \& 7A047QGI) but only one of them arrived.\\[2pt]
    So I need to cancel 4ZM7QGRT and return order 7A047QGI. I started the return online (return id 77436508) but there was no returns label in the box. Can you email me one?\\[2pt]
    \#\#\#\#\#\#\#\#\\[2pt]
    Write an email explaining what the user wants to do, based on the JSON. Include all of the values from the JSON.\\[8pt]
    {[[JSON:]]}\\[2pt]
    [\{"intent": "Return > Label", "return\_id": "641376"\}, \{"intent": "Return > Reschedule Pickup", "return\_id": "641376"\}]\\[8pt]
    {[[Email:]]}\\[2pt]
    Subject: RE: Your rescheduled pickup is booked [641376]\\[2pt]
    Message: My return pickup got scheduled for tomorrow but I won't be in. Can I reschedule to later this week please? Also I never received a shipping label.\\[2pt]
    \#\#\#\#\#\#\#\#\\[2pt]
    Write an email explaining what the user wants to do, based on the JSON. Include all of the values from the JSON.\\[8pt]
    {[[JSON:]]}\\[2pt]
    [\{"intent": "Return > Label", "return\_id": "47089076843001"\}, \{"intent": "Return > Reschedule Pickup", "return\_id": "47089076843001", "new\_date": "tomorrow"\}]\\[8pt]
    {[[Email:]]}\\[2pt]
    Subject: Return \#47089076843001\\[2pt]
    Message: Hi,\\[2pt]
    Can you please email me the label and reschedule pickup for tomorrow if you don't mind? I would appreciate it!!\\[2pt]
    \#\#\#\#\#\#\#\#\\[2pt]
    Write an informal email explaining what the user wants to do, based on the JSON. Include all of the values from the JSON.\\[8pt]
    {[[JSON:]]}\\[2pt]
    [\{"intent": "Return > Label", "return\_id": "4340106665076"\}, \{"intent": "Return > Reschedule Pickup", "return\_id": "4340106665076"\}]\\[8pt]
    {[[Email:]]}\\[2pt]
    Subject:\\[2pt]
}

\newcommand{\eggptgeneration}{
    Reschedule Pickup and Return Label,\\[2pt]
    Message: Hi,\\[2pt]
    I need to reschedule my pickup for Return \#4340106665076 and I never received the shipping label. Can you please help me out?\\[2pt]
    Thanks,\\[2pt]
    Brittany
}

\newcommand{\egevaltrue}{
    [\\
    \textcolor{teal}{\quad \{\\
        \qquad "intent": "Payroll > Correction > Cancel",\\
        \qquad "employee\_name": "Melanie Duke",\\
        \qquad "check\_number": "48P4EQ3T9",\\
        \qquad "date": "2021-12-31"\\
    \quad \},\\}
    \textcolor{orange}{\quad \{\\
        \qquad "intent": "Payroll > Correction > Amount",\\
        \qquad "employee\_name": "Joseph Perez",\\
        \qquad "amount": "\$21.10",\\
        \qquad "date": "2022-01-11"\\
    \quad \},\\}
    \textcolor{violet}{\quad \{\\
        \qquad "intent": "Payroll > Correction > Amount",\\
        \qquad "employee\_name": "Rebecca Mitchell",\\
        \qquad "amount": "\$56.40"\\
    \quad \}\\}
    ]
}

\newcommand{\egevalgen}{
    [\\
    \textcolor{teal}{\quad \{\\
        \qquad "intent": "Payroll > Correction > Cancel",\\
        \qquad "employee\_name": "Melanie Duke",\\
        \qquad "check\_number": "48P4EQ3T9",\\
        \qquad "date": "2021-12-31"\\
    \quad \},\\}
    \textcolor{violet}{\quad \{\\
        \qquad "intent": "Payroll > Correction > Amount",\\
        \qquad "employee\_name": "Joseph Perez",\\
        \qquad "amount": "\$56.40"\\
    \quad \}\\}
    ]
}

\newcommand{\egsubjectfail}{Request for order amendment}
\newcommand{\egbodyfail}{%
    Hello,\\[2pt]
    Could you please amend the following orders:\\[2pt]
    HDJWLRNEP\\[2pt]
    change the price of product 42252271 to 198.81\\[2pt]
    CWXPET\\[2pt]
    change the price of product EKWAUCYM278J to 169.84\\[2pt]
    change the date to 9/21\\[2pt]
    9165483657797\\[2pt]
    change the date to 10/23\\[2pt]
    Thank you
}
\newcommand{\egmessagefail}{
    \textbf{Subject:} \egsubjectfail\\[2pt]
    \textbf{Message:} \egbodyfail
}

\newcommand{\egfailtrue}{
    [\\
        \quad \{\\
            \qquad "intent": "Order > Amendment > Pricing",\\
            \qquad "order\_number": "HDJWLRNEP",\\
            \qquad "product\_id": "42252271",\\
            \qquad "new\_price": "198.81"\\
        \quad \},\\
        \quad \{\\
            \qquad "intent": "Order > Amendment > Pricing",\\
            \qquad "order\_number": "\textcolor{red}{CWXPET}",\\
            \qquad "product\_id": "EKWAUCYM278J",\\
            \qquad "new\_price": "169.84"\\
        \quad \},\\
        \quad \{\\
            \qquad "intent": "Order > Date Change",\\
            \qquad "order\_number": "CWXPET",\\
            \qquad "new\_date": "9/21"\\
        \quad \},\\
        \quad \{\\
            \qquad "intent": "Order > Date Change",\\
            \qquad "order\_number": "9165483657797",\\
            \qquad "new\_date": "10/23"\\
        \quad \}\\
    ]
}

\newcommand{\egfailgen}{
    [\\
        \quad \{\\
            \qquad "intent": "Order > Amendment > Pricing",\\
            \qquad "order\_number": "HDJWLRNEP",\\
            \qquad "product\_id": "42252271",\\
            \qquad "new\_price": "198.81"\\
        \quad \},\\
        \quad \{\\
            \qquad "intent": "Order > Amendment > Pricing",\\
            \qquad "order\_number": "\textcolor{red}{HDJWLRNEP}",\\
            \qquad "product\_id": "EKWAUCYM278J",\\
            \qquad "new\_price": "169.84"\\
        \quad \},\\
        \quad \{\\
            \qquad "intent": "Order > Date Change",\\
            \qquad "order\_number": "CWXPET",\\
            \qquad "new\_date": "9/21"\\
        \quad \},\\
        \quad \{\\
            \qquad "intent": "Order > Date Change",\\
            \qquad "order\_number": "9165483657797",\\
            \qquad "new\_date": "10/23"\\
        \quad \}\\
    ]
}

\title{Generalized Multiple Intent Conditioned Slot Filling}

\author{
    Harshil Shah$^1$ \quad Arthur Wilcke$^1$ \quad Marius Cobzarenco$^1$ \quad \\ 
    {\bf Cristi Cobzarenco$^1$ \quad Edward Challis$^1$ \quad David Barber$^{1,2}$} \\
    $^1$UiPath \quad $^2$University College London
}

\begin{document}
\maketitle
\begin{abstract}
Natural language understanding includes the tasks of intent detection (identifying a user's objectives) and slot filling (extracting the entities relevant to those objectives). Prior slot filling methods assume that each intent type cannot occur more than once within a message, however this is often not a valid assumption for real-world settings. In this work, we generalize slot filling by removing the constraint of unique intents in a message. We cast this as a JSON generation task and approach it using a language model. We create a pre-training dataset by combining DBpedia and existing slot filling datasets that we convert for JSON generation. We also generate an in-domain dataset using GPT-3. We train T5 models for this task (with and without exemplars in the prompt) and find that both training datasets improve performance, and that the model is able to generalize to intent types not seen during training.
\end{abstract}

\section{Introduction}

Natural language understanding (NLU) describes a model's ability to understand the semantic meaning of a sequence of text. When applied to communications data, NLU includes the tasks of intent detection (identifying a user's objectives) and slot filling (extracting the entities relevant to those objectives) \citep{tur-2011-slu}.

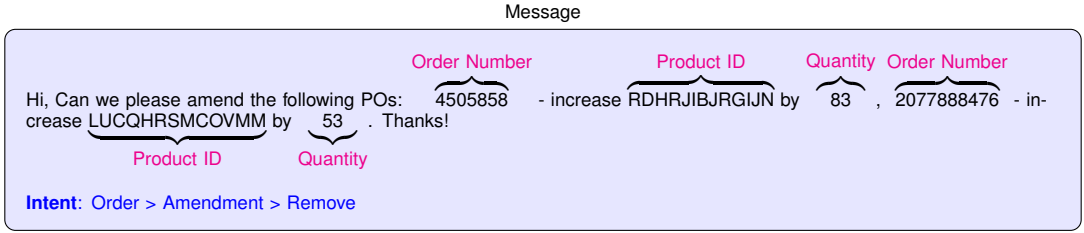
\begin{figure*}[t]
    \centering
    \sffamily
    \begin{tikzpicture}
        \scriptsize
        \node[textbox_input, text width=0.85\linewidth, label=above:Message] (egseqmessage) {\egseqmessage};
    \end{tikzpicture}
    \caption{An example of a message with repeated occurrences of the same intent type. With prior intent detection and slot filling approaches, we can infer which tokens refer to the entities and that (in this example) all of the entities belong to the \texttt{``Order > Amendment > Remove''} intent, but we cannot infer how many times the intent occurs or which Order Numbers should be grouped with which Product IDs and Quantities.}
    \label{fig:intro:example}
\end{figure*}

Slot filling methods assume that the same intent type cannot occur more than once within a given message. However this assumption is often not valid --- for example, the message in Figure \ref{fig:intro:example} contains multiple instances of the user wanting to remove an item from an order. As shown in Figure \ref{fig:intro:example}, the sequence tagging style approaches typically used for slot filling do not generalize when the same intent type occurs multiple times. This makes existing models unusable in many real-world scenarios.

In this work, we generalize slot filling by removing the constraint of unique intents in a message. We do so by treating this as a text generation task where the goal is to generate a JSON string containing the values of the entities grouped by their intents (see Figure \ref{fig:task:example} for an example).

We approach this task using a language model --- an area of research in which we have recently seen remarkable progress \citep{ouyang-2022-instructgpt, schulman-2022-chatgpt}. However to use language models in real-world settings, we often require them to produce output which obeys a pre-specified structure and to not hallucinate (i.e. produce output which is not relevant to the input). To this end, we construct a large pre-training dataset to train the model to extract entities from text and return them in JSON format. 

After pre-training, we find that the model always outputs valid JSON and does not hallucinate (i.e. it only returns values which occur in the original message). However, the pre-training dataset is not sufficient for the model to work well on communications data. To address this, we also construct an in-domain dataset. Communications data is notoriously private therefore we construct a synthetic dataset using GPT-3 \citep{brown-gpt3-2020}, containing over 10K email-JSON pairs with 20 different intent types.

We train T5 models \citep{raffel-t5-2020, chung-flan-t5-2022} with and without exemplars in the prompt, finding that both training datasets improve performance and that the model is able to generalize to intent types not seen during training. We release our code, data and models.\footnote{\url{https://github.com/reinfer/json-extraction-paper}}

\section{Task}
\label{sec:task}

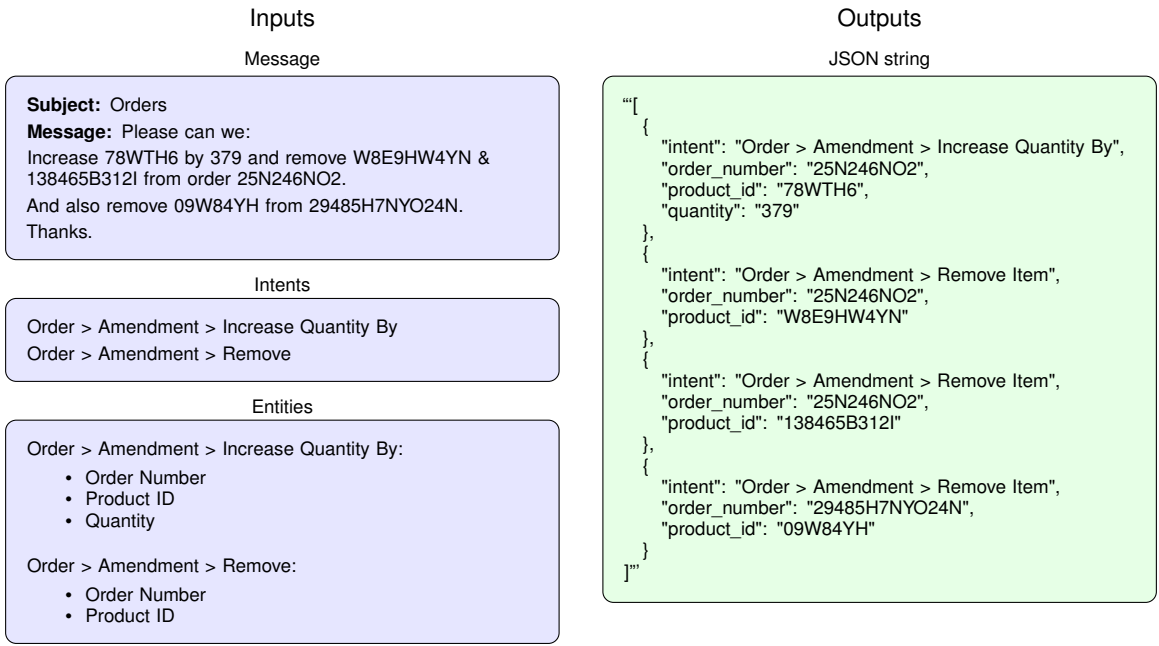
\begin{figure*}
    \centering
    \sffamily
    \begin{tikzpicture}
        \scriptsize
        \node[textbox_input, text width=0.42\linewidth, anchor=north east, shift=({-8pt, 0pt}), label=above:Message] (egmessage) at (0, 0) {\egmessage};
        \node[textbox_input, text width=0.42\linewidth, below=0.5 of egmessage, label=above:Intents] (egintents) {\egintents};
        \node[textbox_input, text width=0.42\linewidth, below=0.5 of egintents, label=above:Entities] (egentities) {\egentities};
        \node[above=0.5 of egmessage] {\footnotesize Inputs};
        \node[textbox_output, text width=0.42\linewidth, anchor=north west, shift=({8pt, 0pt}), label=above:JSON string] (egextraction) at (0, 0) {\egextraction};
        \node[above=0.5 of egextraction] {\footnotesize Outputs};
    \end{tikzpicture}
    \caption{An example of generalized multiple intent conditioned slot filling --- the goal is to generate a JSON string with the values of the given entities from the message. The message, intents and entity names will be used to write a prompt that will be fed as input to a language model (see Figure \ref{fig:prompt}), which will output the JSON string.}
    \label{fig:task:example}
\end{figure*}

Suppose we have a dataset of messages where each message expresses one or more intents from a predefined list. For each intent, we also have a predefined set of named entities to extract. Then, given a message, the set of intents expressed by that message and the named entities to extract for each of those intents, the task is to return the values of the entities, grouped by their intents, from the message. An example is shown in Figure \ref{fig:task:example}.

We assume that the intents for each message are either provided by a labeler or predicted by a separate model. The entities are considered optional --- for a given intent, not every entity will necessarily occur in every message containing that intent.

We treat this as a text generation task where, as per Figure \ref{fig:task:example}, the goal is to generate a JSON string containing the values of the entities\footnote{We use the terms `entity' and `slot' interchangeably throughout.} grouped by their intents.

Specifically, we would like to generate a single array (denoted by square brackets) containing a set of objects (each denoted by curly braces), where each object represents an occurrence of an intent. Each object is a collection of key-value pairs --- these are the entities related to that intent.

\section{Approach}

\subsection{Model}

\begin{figure*}[t]
    \centering
    \sffamily
    \begin{tikzpicture}
        \scriptsize
        \node[textbox_prompt, text width=0.8\linewidth] {\egpromptoneshot};
    \end{tikzpicture}
    \caption{The prompt corresponding to the example in Figure \ref{fig:task:example}. In zero-shot prompting, we only use the part in black (i.e. no exemplars of the completed task are given). In one-shot prompting, we prepend a completed exemplar (shown in \textcolor{purple}{purple}) to the zero-shot prompt to demonstrate how the task is done.}
    \label{fig:prompt}
\end{figure*}

We approach this text generation task by writing a prompt and feeding it to a language model which then generates the JSON string. For the language model we use Flan-T5 \citep{chung-flan-t5-2022}, a variant of the T5 model \citep{raffel-t5-2020} that is tuned to follow instructions in the prompt. The T5 is an encoder-decoder model; we choose it because it performs well across a number of tasks even with a relatively small number of parameters (compared to decoder-only models).

A key benefit of this approach over one-hot encoding the intents and entities (as per the sequence tagging methods) is that the language model is able to leverage the semantic information encoded in the intent and entity names.

\subsubsection{Writing the prompt}
\label{sec:approach:model:prompt}

We structure the prompt as follows: \begin{enumerate}
    \item The instruction.
    \item The intents and entities to extract.
    \item The message.
    \item The opening brackets for the JSON string.
\end{enumerate}

As described above, the prompt is `zero-shot', i.e. it does not include exemplars of the completed task. However, it has been shown that few-shot prompting typically outperforms zero-shot prompting on various tasks \citep{brown-gpt3-2020, chung-flan-t5-2022}. Therefore we also experiment with one-shot prompts, where we put a single exemplar of the completed task in the prompt. Given that our work focuses on relatively long forms of communication (e.g. email), we limit our experiments to only a single exemplar in the prompt (because using more caused out-of-memory problems).

Figure \ref{fig:prompt} shows both the zero-shot and one-shot prompts corresponding to the email from Figure \ref{fig:task:example}, where a completed exemplar is prepended to the zero-shot prompt to demonstrate how the task is done. We use snake-cased entity names as we found this to perform better than title-cased.

\subsection{Data}

\subsubsection{Pre-training}

The Flan-T5 model released by \citet{chung-flan-t5-2022} does not have the ability to generate valid JSON strings and we therefore need to train it to do so.

\paragraph{DBpedia}

We construct a pre-training dataset using DBpedia \citep{Lehmann2015DBpediaA}, a database which contains the abstracts and structured information (e.g. the information boxes often found on the right-hand side) from Wikipedia pages.

Whilst Wikipedia is a different domain compared to communications data, extracting entities from articles is similar to the task described in Section \ref{sec:task}. In particular, Wikipedia articles do not contain intents --- instead, each article has a `type' (e.g. ``City'', ``Film'', ``Politician'', etc.) which we use as a proxy for the intent.

To construct the dataset, we go through each article in the DBpedia database and collect the abstract, the article type, and the infobox properties (i.e. entities). We retain only those entities whose value occurs within the corresponding abstract. In addition, to create the predefined set of entities for each article type, we only retain those that occur in at least 20\% of articles for that type.

In order to create article-JSON pairs which closely resemble the multi-intent communications example in Figure \ref{fig:task:example}, we concatenate together abstracts which contain hyperlinks to each other, and their corresponding entities. An example is given in Figure \ref{wiki:example:full} in Appendix \ref{sec:appendix:pre_training}.

\paragraph{Slot filling datasets}

In addition to the DBpedia dataset, we also convert traditional slot filling datasets into our JSON generation format. For this, we use the ATIS \citep{hemphill-etal-1990-atis}, SNIPS \citep{Coucke2018SnipsVP} and SLURP \citep{bastianelli-etal-2020-slurp} datasets. 

For each of these datasets, we randomly concatenate 2--4 of the utterances and their corresponding intents and entities. This is similar to the approach taken by \citet{qin-etal-2020-agif} for creating the MixATIS and MixSNIPS datasets. However, we don't enforce the restriction that the concatenated utterances must have different intents (as per our generalization of multiple intent slot filling outlined in Section \ref{sec:task}).

\subsubsection{Domain-specific training}
\label{sec:approach:data:gpt}

After training on DBpedia and the slot filling datasets, we find that the model is able to generate valid JSON but it doesn't perform particularly well on communications data (when prompted using either zero-shot or one-shot prompts as per Figure \ref{fig:prompt}). We therefore construct a domain-specific dataset using the \texttt{text-davinci-002} variant of GPT-3 \citep{brown-gpt3-2020}.

We start by creating a set of intents and a set of entities for each of these intents --- see Appendix \ref{sec:appendix:gpt:intents} for a full list. Based on these intents and entities, we create a small set of approximately 100 handcrafted email-JSON pairs (similar to the example in Figure \ref{fig:task:example}). 

Then for each new example we want to generate using GPT-3, we first randomly choose some intents from the predefined list, and for the corresponding entities, we generate realistic but fake values \citep{Faraglia_Faker}. We put these intents and entities into a JSON string and choose 3 email-JSON pairs from the aforementioned handcrafted set such that each of the chosen pairs shares at least one intent with the created JSON. We use these 3 examples to write a prompt which is fed to GPT-3 in order to generate an email which matches the created JSON --- an example prompt and corresponding generation are shown in Figure \ref{fig:gptprompt:example} in Appendix \ref{sec:appendix:gpt:prompt}.

In order to generate more varied examples, we also insert a randomly chosen adjective (e.g. ``informal'', ``tabular'', etc.) into the prompt to describe the message. We split the generated email-JSON pairs into training, validation and testing splits with 8.7K, 1.1K and 1.1K examples respectively.

\subsection{Evaluation}

\begin{algorithm}
    \caption{Evaluating generated vs. true JSONs using fuzzy matching}
    \label{algo:eval}
    \begin{algorithmic}
        \STATE pair the objects from the true and generated arrays which are exact matches
        \\[4pt]
        \FORALL{
            exactly matching pairs
        } \STATE {
            object score: \begin{itemize}[topsep=0pt]
                \item count the pair as a true positive \\[2pt]
            \end{itemize}
            key-value score: \begin{itemize}[topsep=0pt]
                \item count all key-value pairs as true positives (since they all match)
            \end{itemize}
        } \ENDFOR
        \\[8pt]
        \STATE for the remaining (non-exact matching) objects in the true and generated arrays, compute the pairing that maximizes the pairwise fuzzy matching scores between the string representations of the objects
        \\[4pt]
        \FORALL{
            non-exactly matching pairs
        } \STATE {
            object score: \begin{itemize}[topsep=0pt]
                \item count the pair as a false positive \\[2pt]
            \end{itemize}
            key-value score: \begin{itemize}[topsep=0pt]
                \item count the matching key-value pairs as true positives
                \item where the keys match but the values don't, count these as false positives
                \item for any keys that are in the generated object but not in the true one, count these as false positives
                \item for any keys that are in the true object but not in the generated one, count these as false negatives
            \end{itemize}
        } \ENDFOR
        \\[8pt]
        \FORALL{
            remaining (unpaired) objects in the generated array
        } \STATE {
            object score: \begin{itemize}[topsep=0pt]
                \item count the pair as a false positive \\[2pt]
            \end{itemize}
            key-value score: \begin{itemize}[topsep=0pt]
                \item count all key-value pairs as false positives (since this is an `extra' generated object) 
            \end{itemize}
        } \ENDFOR
        \\[8pt]
        \FORALL{
            remaining (unpaired) objects in the true array
        } \STATE {
            object score: \begin{itemize}[topsep=0pt]
                \item count the pair as a false negative \\[2pt]
            \end{itemize}
            key-value score: \begin{itemize}[topsep=0pt]
                \item count all key-value pairs as false negatives (since this is an `extra' true object) 
            \end{itemize}
        } \ENDFOR
    \end{algorithmic}
\end{algorithm}

\begin{figure*}[t]
    \centering
    \sffamily
    \begin{tikzpicture}
        \scriptsize
        \node[textbox_output, text width=0.35\linewidth, anchor=north east, shift=({-8pt, 0pt}), label=above:True] (true) at (0, 0) {\egevaltrue};
        \node[textbox_generation, text width=0.35\linewidth, anchor=north west, shift=({8pt, 0pt}), label=above:Generation] (gen) at (0, 0) {\egevalgen};
    \end{tikzpicture}
    \caption{Using fuzzy matching to compute object level and key-value level F1 scores. First, the exact matches (\textcolor{teal}{teal}) are paired together and then fuzzy matching pairs the \textcolor{violet}{violet} objects. The \textcolor{orange}{orange} object in the true list is left unmatched.\\[8pt]
    \begin{tabular}{l @{\hskip 20pt} l}
        Objects (precision = 0.5; recall = 0.5; F1 = 0.5) & Key-value pairs (precision = 0.86; recall = 0.6; F1 = 0.71) \\
        \quad -- True positives = 1 (teal) & \quad -- True positives = 6 (4 teal + 2 violet) \\
        \quad -- False positives = 1 (violet) & \quad -- False positives = 1 (violet) \\
        \quad -- False negatives = 1 (orange) & \quad -- False negatives = 4 (orange)
    \end{tabular}
    }
    \label{fig:eval:example}
\end{figure*}

As explained in Section \ref{sec:task}, for a given message the true and predicted JSONs each consist of a single array containing a set of objects (one for each occurrence of an intent) where each object is a collection of key-value pairs representing the entities for that intent. When evaluating the model's performance on each message, we are ultimately interested in how many of the objects it is able to correctly generate. However it can also be useful to take a more granular view by looking at how many of the key-value pairs the model correctly generates (i.e. even if an entire object is not correct, it is better to get more of its key-value pairs correct). We therefore compute object level and key-value level F1 scores.

When evaluating a generated array of objects against the true array, it isn't immediately obvious how to pair up the objects from each array in order to compare (especially because the generated objects may not be in the same order as the true array). We approach this problem using fuzzy matching \citep{bachmann_rapidfuzz_2022}; our method to compute the true and false positives and false negatives required for the F1 scores given in Algorithm \ref{algo:eval} and a full example is provided in Figure \ref{fig:eval:example}.

\section{Related work}

The most closely related task to ours in the natural language processing literature is slot filling which, along with intent detection, is used for spoken language understanding (SLU) \citep{tur-2011-slu}.

While we concentrate specifically on slot filling, many SLU methods perform intent detection and slot filling jointly. In this work we assume that the intents are provided by a separate labeling model because performing joint intent detection and slot filling would involve providing all possible intents to the language model --- this would make the prompts extremely long and significantly increase the memory footprint.

Traditionally, slot filling research assumed that each message contained only one intent (and as a result, each token could belong to at most one entity/slot). Slot filling is mostly treated as a sequence tagging task (similarly to named entity recognition) and has been approached using conditional random fields (CRFs) \citep{raymond-2007-crf}, recurrent neural networks (RNNs) \citep{mesnil-2015-rnn, wang-etal-2018-bi}, convolutional neural networks (CNNs) \citep{xu-2013-cnn} and even large language models \citep{chen-2019-bert} such as BERT \citep{devlin-etal-2019-bert}.

In an approach more similar to ours, \citet{fitzgerald-2022-massive} use the mT5 model \citep{xue-etal-2021-mt5}, where the prompt is \texttt{``Annotate:''} followed by the unlabeled message. However their work is still based on each message containing only one intent and (as per the sequence tagging methods) the output is a sequence of labels for each token followed by the intent of the message.

Recently, \citet{gangadharaiah-narayanaswamy-2019-joint} present multiple intent slot filling --- a generalization of traditional slot filling where instead of each message having only one intent, it can have multiple (however they all have to be unique). They approach this task by combining a bi-directional LSTM \citep{hochreiter-1997-lstm} with an attention mechanism and predict which intents each slot belongs to. \citet{qin-etal-2020-agif} and \citet{qin-etal-2021-gl} follow up on this work by replacing the LSTM component with graph interaction frameworks, with the latter achieving state of the art performance.

As described in Section \ref{sec:task}, our proposed task is a further generalization of multiple intent slot filling because we allow the same intent type to occur multiple times in a single message. While the assumption of unique intents in a message may be sensible for shorter forms of communication (e.g. voice commands), it is unrealistic for longer messages (e.g. emails) which may contain features such as tables with each row representing an occurrence of the same intent.

\section{Experiments}

We train two models, one with zero-shot prompts and one with one-shot prompts. The exemplar for the one-shot prompted model is randomly selected from the corresponding training/validation/test set such that the exemplar shares an article type (pre-training) or intent (in-domain training) with the chosen data point (see Figure \ref{fig:prompt} for an example and Section \ref{sec:approach:model:prompt} for more detail on the prompt).

Throughout, we use the \texttt{large} (770M parameter) variant of the Flan-T5 model, initializing the weights at the checkpoint provided by \citet{chung-flan-t5-2022}. We implement our experiments using the PyTorch \citep{paszke-2019-pytorch} and Hugging Face Transformers \citep{wolf-etal-2020-transformers} libraries.

\begin{table}[t]
    \centering
    \begin{tabular}{lcc}
        \toprule
        Prompts & \# GPUs & \# hours \\
        \midrule
        Zero shot & & \\
        \quad pre-training & 2 & 65 \\
        \quad in-domain training & 2 & 2 \\
        One shot & & \\
        \quad pre-training & 4 & 68 \\
        \quad in-domain training & 4 & 2 \\
        \bottomrule
    \end{tabular}
    \caption{The computational budget used for training the models. We used A100 GPUs.}
    \label{tab:train:comp}
\end{table}

\begin{table*}[t]
    \centering
    \begin{tabular}{lcccccc}
        \toprule
         & \multicolumn{3}{c}{Object level} & \multicolumn{3}{c}{Key-value level} \\
        Prompts & Precision & Recall & F1 & Precision & Recall & F1 \\
        \midrule
        Zero-shot & 88.08 & 99.44 & 93.42 & 91.61 & 100.00 & 95.62 \\
        One-shot  & 87.60 & 99.25 & 93.06 & 90.67 & 100.00 & 95.11 \\
        \bottomrule
    \end{tabular}
    \caption{Final performance on the test set of the in-domain data after training the model first on the pre-training data and then on the in-domain training data.}
    \label{tab:results:final}
\end{table*}

\subsection{Training}

We train the model using the standard language modeling maximum likelihood objective.

We first train on the pre-training dataset (DBpedia + slot filling datasets) for 400K iterations, using the Adafactor optimizer \citep{shazeer-2018-adafactor} with a learning rate of $10^{-5}$ and a batch size of 8.

We then further train on the training set of the in-domain data (GPT-3 generations) for 10K iterations, with the same optimization settings as for the first round of training. We measure the log likelihood on the validation set after every 200 iterations and take the final model as the one with the lowest validation log likelihood. When generating outputs from the trained model, we use beam search with a beam size of 3.

The computational budget used for training the models is shown in Table \ref{tab:train:comp}.

\subsubsection{Data augmentation}

When evaluating the model, we found that it would sometimes generate entity-value pairs for entities which weren't requested in the prompt. This is due to the message containing tokens which are similar to entities defined for different intents.

To mitigate this, during training we augment the data by randomly dropping and/or reordering the entities both in the prompt and correspondingly in the target output string.

\subsection{Results}

Table \ref{tab:results:final} shows the precision, recall and F1 scores of the zero-shot and one-shot prompted models on the test set of the in-domain data. Interestingly, the performance is actually slightly better when not showing an exemplar in the prompt. We believe this is because we are evaluating on data which is generated from the same distribution as the training data --- if we were to evaluate on out-of-distribution data, having an exemplar in the prompt is more likely to be useful (see Section \ref{sec:limitations}).

\paragraph{Failure modes} When the model does fail, it tends to be on examples where the target JSON is relatively long and/or where the model gets confused between several ID-like fields; see Figure \ref{fig:fails:example} in Appendix \ref{sec:appendix:fails} for an example.

\subsubsection{Pre-training vs. in-domain training}

\begin{table*}[t]
    \centering
    \begin{tabular}{lcccccc}
        \toprule
         & \multicolumn{3}{c}{Object level} & \multicolumn{3}{c}{Key-value level} \\
        Prompts & Precision & Recall & F1 & Precision & Recall & F1 \\
        \midrule
        Zero-shot &  &  &  &  &  &  \\
        \quad pre-training only                 &  0.00 &  0.00 &  0.00 & 24.70 &  98.40 & 39.49 \\
        \quad in-domain training only           & 81.89 & 97.26 & 88.91 & 86.38 &  99.85 & 92.63 \\
        \quad pre-training + in-domain training & 88.08 & 99.44 & 93.42 & 91.61 & 100.00 & 95.62 \\
        One-shot &  &  &  &  &  &  \\
        \quad pre-training only                 &  0.00 &  0.00 &  0.00 & 25.36 &  97.84 & 40.28 \\
        \quad in-domain training only           & 81.94 & 94.61 & 87.82 & 86.90 &  99.84 & 92.93 \\
        \quad pre-training + in-domain training & 87.60 & 99.25 & 93.06 & 90.67 & 100.00 & 95.11 \\
        \bottomrule
    \end{tabular}
    \caption{Performance on the test set of the in-domain data after the different rounds of training.}
    \label{tab:results:training_rounds}
\end{table*}

Table \ref{tab:results:training_rounds} shows the performance on the in-domain test set after the different rounds of training. After pre-training only, although the model always outputs valid JSON, it performs very poorly at extracting the correct entities on the in-domain data. Note that in this case, the one-shot prompted model performs slightly better than the zero-shot version.

After in-domain training only (i.e. taking the off-the-shelf model and training directly on the in-domain data, with no pre-training), the model is able to perform quite well however we find that doing both pre-training and in-domain training provides the best results. Note that we are unable to directly evaluate the off-the-shelf model without any training, because the curly brace characters (\texttt{`\{'}, \texttt{`\}'}) required for the JSON output are not included in the original vocabulary.




Overall, these results show the benefit of the pre-training process --- even though the pre-training data is semantically different from the in-domain data, it trains the model to be able to extract relevant entities from the input text and output them as JSON, which is then useful for the transfer to the downstream in-domain task.

\subsubsection{Generalization}

To better evaluate how well the model generalizes to new data, we perform a modified version of in-domain training --- we move all of the examples with a \texttt{``Payroll > \dots''} intent from the training and validation sets to the test set, and drop all other non \texttt{``Payroll > \dots''} examples from the test set. This allows us to assess how well the model generalizes to intents not seen during training.

\begin{table*}[t]
    \centering
    \begin{tabular}{lcccccc}
        \toprule
         & \multicolumn{3}{c}{Object level} & \multicolumn{3}{c}{Key-value level} \\
        Prompts & Precision & Recall & F1 & Precision & Recall & F1 \\
        \midrule
        Zero-shot & 77.61 & 93.90 & 84.98 & 82.89 & 100.00 & 90.65 \\
        One-shot  & 74.65 & 91.72 & 82.31 & 80.01 & 100.00 & 88.89 \\
        \bottomrule
    \end{tabular}
    \caption{To better assess the model's ability to generalize to new data, we evaluate its performance on \texttt{``Payroll > \dots''} intents after performing training and validation on only non \texttt{``Payroll > \dots''} examples.}
    \label{tab:results:generalization}
\end{table*}

The results are shown in Table \ref{tab:results:generalization}; they are slightly worse than when all the test intents are also seen during training, but still suggest that the model is able to generalize to new intents. Again, we find that the zero-shot model outperforms the one-shot model. This is even more surprising than the results in Table \ref{tab:results:final} --- having not seen any \texttt{``Payroll > \dots''} examples during training, we would have expected the exemplars in the prompt to be more helpful at test time. We hypothesize that this is because the training and test examples are still very similar in style, even though the \texttt{``Payroll > \dots''} examples are not seen during training.

\section{Limitations}
\label{sec:limitations}

Large language models are computationally expensive to run --- Table \ref{tab:train:comp} shows that the one-shot prompted model takes 280 GPU-hours on A100 GPUs to train. However, the majority of this time is spent pre-training; after this is completed, domain-specific fine-tuning is relatively fast.

Generating the output as a string exposes our method to the risk of hallucination \citep{ji-2022-hallucination}, where the model may output nonsensical, biased or even offensive text unrelated to the input. However since the model has been trained to only ever output text which occurs in the prompt itself (as well as to ignore entities where an appropriate value cannot be found), we believe this risk to be minimal. As a further safeguard, the JSON output can be post-processed to remove any entity-value pairs where either the entity was not requested or the value does not exist in the original message.

Communications data is notoriously private --- as a result, the only in-domain data we have been able to train and evaluate on is synthetic (generated using GPT-3 as per Section \ref{sec:approach:data:gpt}). While we would hope that the one-shot prompted model in particular generalizes to `real' data (because the exemplar in the prompt should help with this adaptation), there is the risk that, in order to use the model on data from a different distribution to the synthetic examples, further fine-tuning may be required.

\section{Conclusion}

In this work, we generalize slot filling by removing the constraint of unique intents in a message. We cast this as a JSON generation task and approach it using a language model. We create a pre-training dataset by combining DBpedia and existing slot filling datasets that we convert for JSON generation, and we generate an in-domain dataset using GPT-3. We train T5 models on these datasets, with and without exemplars in the prompt, finding that both training datasets improve performance and that the model is able to generalize to intent types not seen during training.

This more general approach opens up slot filling to be deployed on a much larger variety of communication types. Furthermore, our pre-training dataset can be used to train models for the more general task of extracting and grouping entities from unstructured text. Our work shows that it is possible to train language models to generate outputs which obey a specific structure --- this type of approach could have applications in several areas, particularly where a language model's outputs are consumed by downstream APIs.

\bibliography{custom}
\bibliographystyle{acl_natbib}

\clearpage

\appendix

\section{Pre-training data}
\label{sec:appendix:pre_training}

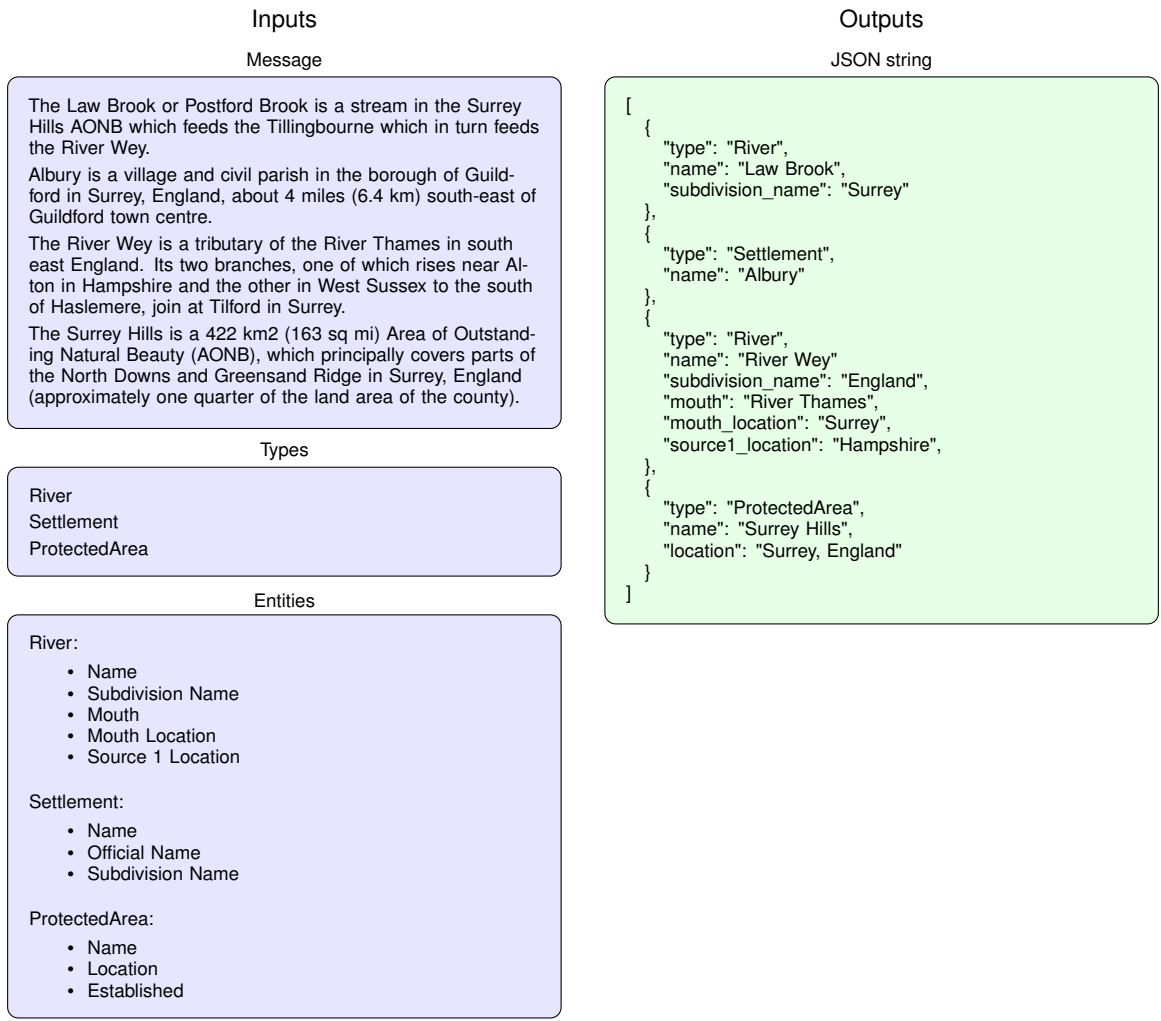
\begin{figure*}
    \centering
    \sffamily
    \begin{tikzpicture}
        \scriptsize
        \node[textbox_input, text width=0.42\linewidth, anchor=north east, shift=({-8pt, 0pt}), label=above:Message] (egwikifull) at (0, 0) {\egwikifull};
        \node[textbox_input, text width=0.42\linewidth, below=0.5 of egwikifull, label=above:Types] (egwikitypes) {\egwikitypes};
        \node[textbox_input, text width=0.42\linewidth, below=0.5 of egwikitypes, label=above:Entities] (egwikientities) {\egwikientities};
        \node[above=0.5 of egmessage] {\footnotesize Inputs};
        \node[textbox_output, text width=0.42\linewidth, anchor=north west, shift=({8pt, 0pt}), label=above:JSON string] (egwikifullextraction) at (0, 0) {\egwikifullextraction};
        \node[above=0.5 of egextraction] {\footnotesize Outputs};
    \end{tikzpicture}
    \caption{An example from the DBpedia pre-training dataset.}
    \label{wiki:example:full}
\end{figure*}

Figure \ref{wiki:example:full} shows a an example from our DBpedia pre-training dataset.

\section{Generating data with GPT-3}
\label{sec:appendix:gpt}

\subsection{Intents}
\label{sec:appendix:gpt:intents}

Below is the full list of intents and entities contained in our in-domain data generated using GPT-3.

\begin{itemize}
    \ttfamily
    \item Order > Cancel
    \begin{itemize}
        \item Order Number
    \end{itemize}

    \item Order > Status
    \begin{itemize}
        \item Order Number
    \end{itemize}

    \item Order > Date Change
    \begin{itemize}
        \item Order Number
        \item New Date
    \end{itemize}

    \item Order > Amendment > Remove Item
    \begin{itemize}
        \item Order Number
        \item Product ID
    \end{itemize}

    \item Order > Amendment > Reduce Quantity By
    \begin{itemize}
        \item Order Number
        \item Product ID
        \item Quantity
    \end{itemize}

    \item Order > Amendment > Increase Quantity By
    \begin{itemize}
        \item Order Number
        \item Product ID
        \item Quantity
    \end{itemize}

    \item Order > Amendment > Change Quantity To
    \begin{itemize}
        \item Order Number
        \item Product ID
        \item Quantity
    \end{itemize}

    \item Order > Amendment > Pricing
    \begin{itemize}
        \item Order Number
        \item Product ID
        \item New Price
    \end{itemize}

    \item Order > Shortage
    \begin{itemize}
        \item Order Number
        \item Product ID
        \item Quantity
    \end{itemize}

    \item Payroll > Correction > Cancel
    \begin{itemize}
        \item Employee Name
        \item Check Number
        \item Amount
        \item Date
    \end{itemize}

    \item Payroll > Correction > Amount
    \begin{itemize}
        \item Employee Name
        \item Amount
        \item Date
    \end{itemize}

    \item Payroll > Employee > Add
    \begin{itemize}
        \item Employee Name
    \end{itemize}

    \item Payroll > Employee > Remove
    \begin{itemize}
        \item Employee Name
    \end{itemize}

    \item Policy > Cancel
    \begin{itemize}
        \item Policy Number
        \item Effective Date
    \end{itemize}

    \item Policy > Change Name
    \begin{itemize}
        \item Policy Number
        \item New Name
        \item Effective Date
    \end{itemize}

    \item Policy > Change Address
    \begin{itemize}
        \item Policy Number
        \item New Address
        \item Effective Date
    \end{itemize}

    \item Product > Availability
    \begin{itemize}
        \item Product ID
        \item Size
    \end{itemize}

    \item Product > Measurements
    \begin{itemize}
        \item Product ID
        \item Size
    \end{itemize}

    \item Return > Label
    \begin{itemize}
        \item Return ID
    \end{itemize}

    \item Return > Reschedule Pickup
    \begin{itemize}
        \item Return ID
        \item New Date
    \end{itemize}
\end{itemize}

\subsection{Prompt}
\label{sec:appendix:gpt:prompt}

\begin{figure*}[t]
    \centering
    \sffamily
    \begin{tikzpicture}
        \scriptsize
        \node[textbox_prompt, text width=0.8\linewidth, label=above:Prompt] (eggptprompt) {\eggptprompt};
        \node[textbox_generation, text width=0.8\linewidth, label=above:Generation, below=0.6 of eggptprompt] (eggptgeneration) {\eggptgeneration};
    \end{tikzpicture}
    \caption{Prompting GPT-3 to generate an email which matches the given JSON.}
    \label{fig:gptprompt:example}
\end{figure*}

Figure \ref{fig:gptprompt:example} shows an example prompt and corresponding generation for creating in-domain data using GPT-3.

\section{Failure modes}
\label{sec:appendix:fails}

\begin{figure*}[t]
    \centering
    \sffamily
    \begin{tikzpicture}
        \scriptsize
        \node[textbox_input, text width=0.8\linewidth, anchor=south, shift=({0pt, 16pt}), label=above:Message] (message) at (0, 0) {\egmessagefail};
        \node[textbox_output, text width=0.42\linewidth, anchor=north east, shift=({-8pt, 0pt}), label=above:True] (true) at (0, 0) {\egfailtrue};
        \node[textbox_generation, text width=0.42\linewidth, anchor=north west, shift=({8pt, 0pt}), label=above:Generation] (gen) at (0, 0) {\egfailgen};
    \end{tikzpicture}
    \caption{An example failure mode --- the model outputs the incorrect order number for the second price amendment.}
    \label{fig:fails:example}
\end{figure*}

When the model does fail, it tends to be on examples where the target JSON is relatively long and/or where the model gets confused between several ID-like fields; see Figure \ref{fig:fails:example} for an example.

\end{document}